\documentclass[letterpaper, 10 pt, conference]{ieeeconf}  
\IEEEoverridecommandlockouts                             
\usepackage{amsmath}
\usepackage{todonotes}
\usepackage{nicefrac}
\usepackage[hidelinks]{hyperref}

\usepackage{cite} % groups citations e.g. [1]-[4]

\newcommand{\myvec}[1]{\boldsymbol{#1}}

\newcommand{\vf}{\myvec{f}}

\newcommand{\vq}{\myvec{q}}

\newcommand{\vu}{\myvec{u}}
\newcommand{\vv}{\myvec{v}}

\newcommand{\vx}{\myvec{x}}

\newcommand{\vA}{\myvec{A}}
\newcommand{\vB}{\myvec{B}}

\newcommand{\vJ}{\myvec{J}}

\title{\LARGE \bf
The Control Toolbox - An Open-Source C++ Library for Robotics, Optimal and Model Predictive Control
}

\author{Markus Giftthaler$^{\dagger}$, Michael Neunert$^{\dagger}$, Markus St\"auble and Jonas Buchli$^{*}$% <-this % stops a space
\thanks{\scriptsize $^\dagger$These authors contributed equally to this work.}
\thanks{\scriptsize $^*$Agile \& Dexterous Robotics Lab, ETH Z\"urich, Switzerland. \{\mbox{mgiftthaler@ethz.ch}, \mbox{neunertm@gmail.com}, \mbox{markusta@ethz.ch}, \mbox{buchlij@ethz.ch}\}}
}

\begin{document}

\maketitle
\thispagestyle{empty}
\pagestyle{empty}

%%%%%%%%%%%%%%%%%%%%%%%%%%%%%%%%%%%%%%%%%%%%%%%%%%%%%%%%%%%%%%%%%%%%%%%%%%%%%%%%
\begin{abstract}
We introduce the Control Toolbox (CT), an open-source C++ library for efficient modeling, control, estimation, trajectory optimization and Model Predictive Control. The CT is applicable to a broad class of dynamic systems but features interfaces to modeling tools specifically designed for robotic applications. 
This paper outlines the general concept of the toolbox, its main building blocks, and highlights selected application examples. 
The library contains several tools to design and evaluate controllers, model dynamical systems and solve optimal control problems.
The CT was designed for intuitive modeling of systems governed by ordinary differential or difference equations. It supports rapid prototyping of cost functions and constraints and provides standard interfaces for different optimal control solvers.
To date, we support Single Shooting, the iterative Linear-Quadratic Regulator, Gauss-Newton Multiple Shooting and classical Direct Multiple Shooting. We provide interfaces to general purpose NLP solvers and Riccati-based linear-quadratic optimal control solvers.
The CT was designed to solve large-scale optimal control and estimation problems efficiently and allows for online control of dynamic systems. Some of the key features to enable fast run-time performance are full compatibility with Automatic Differentiation, derivative code generation, and multi-threading.
Still, the CT is designed as a modular framework whose building blocks can also be used for other control and estimation applications such as inverse dynamics control, extended Kalman filters or kinematic planning.
The CT is available as open-source software under the Apache v2 license and can be retrieved from \\ \url{https://bitbucket.org/adrlab/ct}.
\end{abstract}

%%%%%%%%%%%%%%%%%%%%%%%%%%%%%%%%%%%%%%%%%%%%%%%%%%%%%%%%%%%%%%%%%%%%%%%%%%%%%%%%
\section{Introduction}
\subsection{What is the Control Toolbox?}
A common task for robotics researchers and practitioners is to model systems, implement equations of motion and design model-based controllers, estimators, planning algorithms, etc.
Sooner or later, one is confronted with questions of efficient implementation, computing derivatives, formulating cost functions and constraints or running controllers in a model-predictive control fashion.

The Control Toolbox is specially designed for these tasks. It is written entirely in C++ and has a strong focus on highly efficient code that can be run online (in the loop) on robots or other actuated hardware.
A significant contribution of the CT is its implementation of optimal control algorithms, spanning a range from simple LQR reference implementations to constrained Model Predictive Control.
The CT supports Automatic Differentiation (Auto-Diff) and allows to generate derivative code for arbitrary scalar and vector-valued functions. 
The toolbox was designed with usability in mind, allowing users to apply advanced concepts such as Nonlinear Model Predictive Control (NMPC) or numerical optimal control quickly and with minimal effort.
In contrast to highly integrated frameworks, the CT follows a modular library approach. Thus, it can be easily interfaced with external modeling and solver frameworks. Additionally, building blocks such as Automatic-Differentiation are usable in several applications including optimal control, classical feedback control (e.g. LQRs), Kalman filtering or sensitivity analysis. In summary, several elements of a planning and control pipeline for an application such as joint space PID control, inverse dynamics control and motion planning can be developed using a single library.

The CT has been designed to provide the tools needed for fast development and evaluation of control methods while being optimized for efficiency and allowing for online operation. While the emphasis lies on control, the provided tools can also be used for simulation, estimation or other optimization applications.

There are four key components of the control toolbox: Modelling of continuous and discrete-time dynamical systems, Automatic Differentiation, optimal control algorithms and Rigid Body Dynamics algorithms. For many of these components, we have carefully selected existing implementations and provide a seamless integration between them without creating a rigid framework. Instead, CT offers easy to use tools for fast prototyping, such as numerical integrators or LQR design. More complex approaches are provided in the form of reference implementations, such as generating a whole-body Nonlinear Model Predictive Control setup for a robot based solely on a semantic description.

\subsection{Related Work}
When looking at software tools in robotics, the library that matches the scope of CT the closest is Drake~\cite{tedrake:2014:drake}. However, Drake started out as a Matlab implementation, which made it unsuitable for hard-realtime and online control. While its codebase is gradually moving towards C++, not all features have been ported yet and Auto-Diff support is limited. Another popular software library is MuJoCo~\cite{todorov2012mujoco}, which excels at simulation but follows a closed-source policy.

There are many optimal control toolboxes outside of the robotics community which focus on transcribing and solving nonlinear optimal control problems and influenced the development of CT. Notable examples are ACADO~\cite{houska2011acado}, its successor ACADOS~\cite{acados}, PSOPT~\cite{becerra2010psopt}, the closed-source toolbox MUSCOD~\cite{diehl2001muscod} as well as the commercial tools GPOPS~\cite{patterson2014gpops} and ForcesPro~\cite{forcespro}.
Broadly speaking, there are three categories of solvers frequently used in optimal control: linear-quadratic optimal control solvers such as HPIPM~\cite{frisonalgorithms}, general quadratic programming solvers such as qpOases~\cite{Ferreau2014qpoases}, and  general-purpose nonlinear-programming packages, such as IPOPT~\cite{wachter:2006:ipopt}, SNOPT~\cite{gill:2005:snopt} or NLopt~\cite{nlopt}. Developing such solvers is a research field by itself and not the primary scope of the CT. Therefore, we provide an interface to SNOPT, IPOPT, and HPIPM as well as custom implementations of iLQR~\cite{todorov2005ilqg} and Gauss-Newton Multiple Shooting~\cite{giftthaler2017family}. This gives the user the opportunity to evaluate different solver types and implementations.

For Automatic Differentiation, we rely on CppAD paired with the code-generation framework CppADCodeGen~\cite{bell2012cppad}. For a more detailed review on Auto-Diff frameworks as well as why symbolic differentiation as e.g. available in Matlab~\cite{web:matlab}, Mathematica~\cite{web:mathematica} or MapleSim~\cite{web:maplesoft} compares unfavorably to Auto-Diff regarding speed, we refer to~\cite{giftthaler2017autodiff}.

For modelling rigid body dynamics, there exists a variety of mature libraries, with notable examples being RBDL~\cite{felis2016rbdl}, DART~\cite{dart}, Pinocchio~\cite{pinocchioweb}, iDynTree~\cite{idyntree} and RobCoGen~\cite{frigerio:2016:robcogen}. Due to its modularity, CT can be interfaced with either of these libraries. We decided to provide a reference interface to RobCoGen due to its wide Auto-Diff support and suitability for online control. Support and interfaces to other rigid body dynamics libraries are going to be added to CT once their Auto-Diff support improves.

\subsection{Scope}
For control, especially numerical optimal control in a robotics context, there are many individual libraries available that provide key ingredients such as modeling frameworks, Auto-Diff, and optimal control solvers. However, due to different data representations, modeling assumptions, basic conventions and the lack of reference implementations in a robotics context, integrating these components is a tedious, time-consuming and error-prone process. Also, for researchers entering the field, the correct choice of modeling framework or solver is difficult to make since the solvers' scope and functionality differ strongly. Furthermore, their performance highly depends on the specific modeling implementation.

For this reason, the CT aims at providing users with the tools to quickly implement well-established control methods and combine their problem with different solvers, transcription methods, and modeling approaches. Since the CT is open-source, it is easy to adapt individual components to specific use cases or integrate custom modeling or solver frameworks. On top of this integration, the CT follows a holistic approach to robot control: the individual components are not only useful for numerical optimal control but can also be employed for classical feedback control, inverse dynamics control, estimation and planning. The CT provides features such as Kalman filtering or contact constraint projection for Rigid Body dynamics~\cite{pardo2017projection}. Thus, the CT can be used for several aspects of a robotics control and planning toolchain: from low-level realtime closed-loop control to kinematic planning or dynamic whole-body trajectory optimization.

\subsection{Structure of This Paper}
This paper is structured as follows. 
In Section~\ref{sec:overview}, we present an overview of the CT's design and implementation and give an outline of its structure. The different main modules of the CT are highlighted in Sections~\ref{core_module} to~\ref{models_module}. Selected application examples are given in Section~\ref{sec:application_examples}. For real-time applications, optimizing runtime performance is an important issue, on which we comment in Section~\ref{sec:performance_opt}.
The paper is concluded by important links and licence information (Section~\ref{sec:further_information}) and acknowledgements to additional contributors in Section~\ref{sec:contributors}.

\section{Overview}
\label{sec:overview}
\subsection{Fundamental Dependencies}
The CT is written in C++ and has been tested under Ubuntu 14.04 and 16.04 with library versions as provided in the package sources. Building the CT requires a C++ compiler with C++11 support. 
Since the CT is designed as a toolbox rather than an integrated application, we tried to provide maximum flexibility to the users. Therefore, it is not tied to a specific middleware such as ROS and dependencies are kept at a minimum. 
The two essential dependencies for CT are Eigen~\cite{web:eigen} and kindr~\cite{kindr} (which is based on Eigen). Eigen is a popular library for linear algebra in C++ and provides efficient implementations of standard matrix operations as well as more advanced linear algebra methods. Kindr is a header only kinematics library which builds on top of it and provides data types for different rotation representations such as quaternions, Euler angles or rotation matrices.

\subsection{Structure and Modules of the CT}
The Control Toolbox consists of three main modules. The core module (ct\_core), the optimal 
control module (ct\_optcon) and the rigid body dynamics module (ct\_rbd). 
There is a clear hierarchy between the modules, which means the modules depend on each other in this order. For example, one can use the core module
without ct\_optcon or ct\_rbd.
\begin{itemize}
\item \emph{ct\_core} provides general type definitions and mathematical tools. For example, it contains most data type definitions, definitions for systems and controllers, as well as basic functionality such as numerical integrators for differential equations. 
\item \emph{ct\_optcon} builds on top of the `core' module and adds infrastructure for defining and solving optimal control problems. It contains the functionality for defining cost functions, constraints, solver backends and a generic NMPC wrapper. 
\item \emph{ct\_rbd} provides tools for modelling rigid body dynamics systems and interfaces with ct\_core and ct\_optcon. 
\end{itemize}

For testing as well as for giving examples, we provide a fourth module: the `models' module (ct\_models) contains various robot models including a quadruped, a robotic arm, a normal quadrotor and a quadrotor with a slung load. These four different modules are detailed in Sections~\ref{core_module}-\ref{models_module}.

\section{Core Module}
\label{core_module}
\subsection{Basic System Definitions}
The core module defines basic data types and interfaces to describe non-linear system dynamics of the forms
\begin{align}
\dot{\vx} &= \vf(\vx(t), t)) \label{eq:system}\\
\dot{\vx} &= \vf(\vx(t), \vu(t), t) \ \text{.} \label{eq:controlled_system}
\end{align}
The right-hand side (RHS) of Equation~\eqref{eq:system} only depends on the time and state $\vx(t)$ (\emph{core::StateVector}) and is called a \emph{core::System}.
The RHS of Equation~\eqref{eq:controlled_system} additionally depends on the control input $\vu(t)$ (\emph{core::ControlVector}) and is named \emph{core::ControlledSystem}.
The dynamics equations can be implemented by the user in any desired way but are currently restricted to ordinary ODEs and difference equations. For the remainder of this section, we limit the scope to a continuous-time perspective. Note that for modeling robotic systems in continuous-time, the rigid body dynamics module provides a variety of tools, which are detailed in Section~\ref{sub:rbd_module}.

As the name suggests, the \emph{core::ControlledSystem} provides the interface for closing a feedback control loop.
Every controlled system can take a pointer to a control law deriving from \emph{core::Controller}. Full flexibility for implementing a policy of general form $\vu(\vx(t), t)$ is given to the user. This includes special cases where the control is merely constant, depending on neither $\vx(t)$ nor $t$, time-varying, only depending on $t$, or a general feedback controller, depending on both $\vx(t)$ and t.

We provide a set of pre-defined control laws, which includes a \emph{core::ConstantController} with fixed $\vu$, a classical PID controller (\emph{core::PIDController}), or a full time-varying \emph{core::StateFeedbackController} with feedforward term of form $\mathbf u_{ff}(t) + \mathbf K(t)(\mathbf x(t)- \mathbf x_{ref}(t))$.

\subsection{Integration and Simulation}
The CT provides different numerical integrators (\emph{core::Integrator}). We offer own implementations and integrators based on `boost odeint'~\cite{odeint}.

The CT currently features fixed-step integrators like Euler and fourth-order Runge-Kutta as well as different (error controlled) variable step integrators. Additionally, for symplectic systems (\emph{core::SymplecticSystem}) a semi-implicit Euler integrator (\emph{core::SymplecticIntegrator}) is available, which can help with stiff systems.
All integrators take a pointer to a system and return trajectories (\emph{core::DiscreteTrajectory}), i.e. timed series of states and control inputs (\emph{core::StateTrajectory} and \emph{core::ControlTrajectory}) respectively.
These trajectories can be either equidistant in time or unevenly sampled. 
In both cases, an interpolation strategy can be applied to obtain states and inputs at a specific time which is not directly stored. 

For rapid prototyping and testing of control loops, we provide a \emph{core::ControlSimulator} which allows running controllers and system integration in parallel and in real-time. Please note, however, that the CT cannot replace a high-fidelity physics simulator. For such purposes, we refer for example to~\cite{sherman2011simbody}.

\subsection{Computing Derivatives}
\begin{table}
\centering
\caption{Comparing different options to obtain derivatives}
\label{tab:comparing_derivatives}
\begin{tabular}{|c|c|c|c|c|}
\hline
\textit{Derivative} & \textit{Numerical} & \textit{Computation} & \textit{Setup} & \textit{Error} \\
\textit{method}    & \textit{Accuracy} & \textit{Speed} & \textit{Time} & \textit{Safety} \\
\hline \hline
Num-Diff & $-$ & $-$ & $+++$ & $+++$ \\
Analytic Deriv. & $+++$ & $++$ & $-$ & $-$ \\
Symbolic Engine & $+++$ & $+$ & $+$ & $++$ \\
Auto-diff & $+++$ & $+$ & $++$ & $++$ \\
Auto-diff Codegen & $+++$ & $+++$ & $++$ & $++$ \\
\hline
\end{tabular}
\end{table}

The CT can be used to compute derivatives of arbitrary vector-valued smooth nonlinear functions $\vf(\vx)$. 
For computing first order derivatives (Jacobians) $\vJ = \frac{d \vf}{d \vx}$, the most-widespread methods are
\begin{enumerate}
\item Numerical differentiation, e.g. by the method of finite-differences,
%\begin{equation}A = \frac{\vf(\vx+h, \vu) - f(x)}{h} \end{equation}
\item Analytical derivation, e.g. performed manually,
\item Symbolic math engines,
\item Automatic Differentiation, also known as Algorithmic Differentiation, with an optional source code generation step.
\end{enumerate}

The different approaches are compared in Table~\ref{tab:comparing_derivatives}. 
Automatic Differentiation allows to conveniently obtain derivative information: it relieves the user from computing analytical derivatives manually or symbolically\footnote{Auto-Diff uses graph structures to compute derivatives. Hence, it is inherently different from symbolic engines such as Maple or Maxima.}, which may be intractable for complex systems. However, it is as accurate and fast as analytic derivatives and outperforms numerical differentiation in terms of accuracy and speed while providing a similar level of convenience. 
Combining Automatic Differentiation with source code generation (Auto-Diff Codegen) results in the best runtime of the differentiation methods supported by CT. For a detailed review and numerical examples, the interested reader is referred to~\cite{giftthaler2017autodiff}.

\subsection{Linearizing Dynamic Systems}
The CT defines the structure of a linear system (\emph{core::LinearSystem}) as
\begin{equation}
\dot{\vx}(t) = \vA(t) \vx(t) + \vB(t) \vu(t)
\end{equation}
where $\vA$ and $\vB$ are the Jacobians of a non-linear, time-varying system evaluated at desired setpoints for $\vx$ and $\vu$.
%\begin{align}
%\vA(t) = \frac{d\vf(\vx,\vu)}{d\vx} |_{\vx=\vx_l(t), \vu=\vu_l(t)} \\
%\vB(t) = \frac{d\vf(\vx,\vu)}{d\vu} |_{\vx=\vx_l(t), \vu=\vu_l(t)}
%\end{align}
In order to compute this linearization for a non-linear system, CT provides two different helper classes. The \emph{core::SystemLinearizer} takes a \emph{core::ControlledSystem} and applies numerical differentiation to compute the Jacobians. Alternatively, the \emph{core::AutoDiffLinearizer} can be used to apply Auto-Differentiation for the Jacobians which is more accurate than numerical differentiation. Finally, Auto-Differentiation is combined with code generation in \emph{core::ADCodegenLinearizer} which is as accurate as analytical derivatives and typically fast to evaluate. The code-gen linearizer employs a technique called \emph{just-in time compilation} (JIT), which generates the derivative code at runtime. Since this can take a few seconds, the derivative code can be stored to file and compiled in separate libraries. Examples for this approach are given in ct\_models, see Section~\ref{models_module}.

\subsection{Computing Approximated and Exact Sensitivities}
Many control algorithms, for example the direct approaches to optimal control shown in Section~\ref{sub:optcon_module}, require a discrete-time approximation of the nonlinear system dynamics of form
%
%\begin{equation}
$\vx_{n+1} = \vA_n \vx_n + \vB_n \vu_n$,
%\end{equation}
%
where we call $\vA_n$ and $\vB_n$ `sensitivities'. In many cases it may suffice to approximate these matrices based on the continuous-time counterparts $\vA(t)$, $\vB(t)$ and a simple Forward-Euler, Backward-Euler or Tustin discretization scheme. The CT provides the \emph{core::SensitivityApproximation} class, which can be used to compute such low-order approximations in a straight-forward way.

However, especially when aiming at a coarse time-discretization while dealing with a highly nonlinear dynamic systems, it can be beneficial to use higher-order integration schemes to compute $\vA_n$, $\vB_n$. The \emph{core::SensitivityIntegrator} solves the integrals
\begin{align}
\vA_n &= \int_0^{\Delta t} \frac{\partial \vf(\vx(t+\tau), \vu(t+\tau), t+\tau)}{\partial \vx(t)} d\tau \notag \\
\vB_n &= \int_0^{\Delta t} \frac{\partial \vf(\vx(t+\tau), \vu(t+\tau), t+\tau)}{\partial \vu(t)} d\tau \notag
\end{align}
for a given starting time $t$ and time-step $\Delta t$ by means of integrating a Sensitivity ODE. Special cases for obtaining exact sensitivities for symplectic integration schemes are included, too.
Exact sensitivities can help to robustify and improve the convergence behavior of many optimal control algorithms in the CT, which are summarized below.

\section{Optimal Control Module}
\label{sub:optcon_module}

A broad variety of model-based optimal control tasks can be formulated as continuous-time optimal control problems. From a robotics perspective, this includes tasks such as agile flight, reaching an object in a cluttered scene, moving a mobile manipulator or quadrupedal locomotion. In direct optimal control, the continuous-time optimal control problem is first transcribed into a numerically tractable discrete problem. Two possible ways to complete this step are:
\begin{enumerate}
\item Transcribing the problem into a nonlinear program (NLP) using multiple-shooting, single shooting or direct collocation and subsequently solving it using standard NLP solvers such as IPOPT or SNOPT.
\item Using iterative Riccati-based shooting methods derived from the Principle of Optimality such as DDP~\cite{mayne1966ddp}, their Gauss-Newton counterparts, iLQR~\cite{todorov2005ilqg} or Gaus-Newton Multiple Shooting (GNMS)~\cite{giftthaler2017family}. These methods are popular due to their overall efficiency and linear time complexity.
\end{enumerate}

The package ct\_optcon covers both classical off-the-shelf NLP solvers and custom Riccati-based solutions, paired with different flavors of Single and Multiple Shooting. An important design feature is the CT's modularity, which allows combining different cost functions, dynamics, constraints, and solvers in an almost arbitrary way and therefore allows for rapid prototyping of optimal control setups, including Nonlinear Model Predictive Control.

\subsection{Cost Functions}
\label{sec:cost_functions}
The cost function package provides means of quickly prototyping objective functions based on a highly modular approach. A CT cost function is assumed to consist of a sum of elementary cost function building blocks, which are called `terms'. Each term evaluates to a scalar as a function of the current time, control input, and state and derives from \emph{optcon::TermBase}.

The overall cost function is designed such that it holds intermediate terms and final terms, which can be assigned individually.
The intermediate and final costs are then given as the sums over the evaluations of all intermediate and final terms. Equivalently, the intermediate and final derivatives result as the sums of the individual intermediate and final term gradients. 
The cost function package supports both analytic derivatives for terms as well as Automatic Differentiation and just-in-time compilation (JIT) up to second order derivatives.

We offer a selection of frequently used standard cost function terms, which penalize the deviations from given control and state reference points, including a purely quadratic term (\emph{optcon::TermQuadratic}), a cross-term (\emph{optcon::TermMixed}) and a purely linear term (\emph{optcon::TermLinear}). Furthermore there are terms for tracking reference trajectories in state and control (\emph{optcon::TermQuadTracking}) and terms which formulate soft constraints on state and control variables (\emph{optcon::TermStateBarrier}).

All existing terms can be automatically constructed from text-files, in which the cost function weights and parameters can be structured in a simple manner. For custom terms, reading from a file is simple to implement thanks to a pre-specified set of loading methods. Additionally, all terms can be made-time-varying using time-activation functions, which can be used to introduce way-point costs, for example.

\subsection{Constraints}
The constraint package generalizes the modular idea presented for cost functions in Section~\ref{sec:cost_functions} to vector-valued functions.
The corresponding elementary building blocks derive from \emph{optcon::ConstraintBase} and again support both analytic derivatives, Automatic-Differentiation and Auto-Diff with JIT. For constraints, the terms are not summarized but stacked in a so-called `constraint container'. Every container additionally features an upper and a lower bound. For constraints, we currently only support first-order derivatives (\emph{optcon::LinearConstraintContainer}). To date, the predefined terms include simple linear path inequality constraints and box constraints on states and controls.

\subsection{Optimal Control Problem Containers}
A \emph{optcon::OptConProblem} is a unified container for nonlinear controlled system dynamics, Equation~\eqref{eq:controlled_system}, nonlinear cost functions, nonlinear constraints, a time horizon variable and an initial state. It serves as the main interface between a user and the different implementations of optimal control algorithms and NMPC.

Similarly, the container \emph{optcon::LQOCProblem} is dedicated to constrained linear-quadratic optimal control problems. However, this container is designed to directly store the linearized dynamics, the Jacobians and Hessians of the cost function and the constraint Jacobians in matrix representation.

\subsection{LQR and Linear Quadratic Solvers}
The CT provides C++ code for different variants of the classical Linear Quadratic Regulator. We provide direct and iterative solvers for the continuous-time Algebraic Riccati Equation (\emph{optcon::CARE}), and iterative solvers for the discrete-time Algebraic Riccati Equation (\emph{optcon::DARE}). Those can be used to design infinite-horizon LQR controllers and state- and disturbance estimators in both continuous- and discrete time. Furthermore, there is a time-varying, finite-horizon discrete-time LQR version available (\emph{optcon::FHDTLQR}).

For unconstrained linear-quadratic optimal control problems, the CT offers a custom Riccati solver, \emph{optcon::GNRiccatiSolver}, which achieves high efficiency using advanced options such as fixed Hessian regularization.

For constrained LQ optimal control problems the CT includes an interface to the interior point solver HPIPM~\cite{frisonalgorithms}, which is a competitive solver for constrained optimal control problems: it features linear time complexity thanks to a Riccati factorization and uses a linear algebra package with CPU architecture specific optimization~\cite{blasfeo}.

\subsection{NLP Problems and Solvers}
\label{sec:nlps}
A unified, Eigen-based interface for formulating nonlinear programming problems (\emph{optcon::Nlp}) and solving them (\emph{optcon::NlpSolver}) is part of the CT. To date, we provide interfaces to the free interior-point solver IPOPT~\cite{wachter:2006:ipopt} and the commercial SQP-solver SNOPT~\cite{gill:2005:snopt}.

\subsection{Gauss-Newton Shooting Algorithms with Riccati solvers}
The CT implements a family of Gauss-Newton Multiple Shooting algorithms in both unconstrained and constrained fashion~\cite{giftthaler2017family}. This family of algorithms performs Sequential Quadratic Programming on the original nonlinear optimal control problem, uses appropriate Riccati solvers to solve linear-quadratic sub-problems efficiently, and utilizes a line-search over a merit function for globalization. The algorithms employ a piece-wise constant control parameterization. A famous limit case of the family of algorithms is the iterative Linear Quadratic Regulator (iLQR). The details of these algorithms have been extensively covered elsewhere~\cite{todorov2005ilqg,giftthaler2017family}. However, we note that the CT shows how to integrate these algorithms in a single framework at almost identical computational cost.
These algorithms are particularly powerful for unconstrained problems with long time horizons or very fine control discretizations. Additionally, at every iteration, they design a time-varying state-feedback control law, which generalizes the policy in the vicinity of the optimal solution.

\subsection{Classical Direct Multiple Shooting}
Complementary to GNMS, the CT also implements the original Direct Multiple Shooting (DMS) method by Bock and Plitt~\cite{bock1984direct}, which we solve using a classical NLP solver (see Section~\ref{sec:nlps}. We provide this method separately since it complements the other algorithms in several aspects. 
While GNMS currently only supports a constant control parameterization, DMS also supports linear interpolation. DMS in combination with IPOPT can furthermore leverage exact Hessians or other Hessian approximations.
DMS furthermore supports adaptive step-size integration. Lastly, DMS can make use of more advanced globalization techniques as employed by the NLP solvers, such as complex filter schemes~\cite{nocedal}.
However, for problems with long time horizons, the DMS implementation cannot compete with GNMS or iLQR at runtime, due to computational limitations of the currently available off-the-shelf NLP solvers.

\subsection{Nonlinear Model Predictive Control}
Thanks to a dedicated design of interfaces between solvers and the optimal control problem definition, the CT optimal control problem solvers can be automatically run in Nonlinear Model Predictive Control fashion using the class \emph{optcon::MPC}. The latter offers options like automatic warm-starting, pre-integration for delay-compensation, different modes to handle time horizons (e.g. receding horizon, fixed time horizon) and offers explicit support for real-time iteration schemes~\cite{diehl2005rti}.
For a detailed example of NMPC using a GNMS nonlinear optimal control solver, the reader is referred to the ct\_optcon online tutorial.

\section{Rigid Body Dynamics Module} 
\label{sub:rbd_module}
Generally speaking, the main task of the rigid body dynamics module ct\_rbd is to provide wrappers that map specialized RBD code into a general ordinary differential equation of form~\eqref{eq:controlled_system}, cost functions, and constraints.

The rigid body dynamics module currently relies on RobCoGen~\cite{frigerio:2016:robcogen}, a code-generation framework for rigid body dynamics and kinematics. To model a new robot in the Control Toolbox, an additional code-generation step is required to create the dynamics and kinematics equations based on a user-provided semantic robot description.
To date, RobCoGen is the only C++ rigid body dynamics engine that supports Automatic Differentiation, a major ingredient for our NMPC applications, see e.g.~\cite{neunert2017mpc}.

Generating a new robot model based on the code-generation output of RobCoGen is straight-forward and amounts to creating a single header file with only a few lines of code. Essentially, one needs to specify kinematic branches and end-effector locations.
In the background, ct\_rbd creates containers and wrappers which allow convenient access to the generated robot dynamics and kinematics functions as well as force-transforms and Jacobians.
For fixed-base systems, the dynamics container is the class \emph{rbd::FixBaseFDSystem}, for floating-base systems it is \emph{rbd::FloatingBaseFDSystem}. The floating-base state is
\begin{equation*}
    \vx = [{}_W\vq^\top ~ {}_B\dot{\vq}^\top]^\top = [{}_W \mathbf \Omega^\top_B ~ {}_W\vx^\top_B ~ \mathbf \theta^\top ~ {}_B\mathbf \omega^\top_B ~ {}_B\vv^\top_B ~ \dot{\mathbf{\theta}}^\top]^\top
\end{equation*}
where ${}_W\Omega_B$ and ${}_W\vx_B$ define base orientation and position expressed in the inertial (`world') frame. ${}_B\omega_B$ and ${}_B\vv_B$ represent local angular and linear velocity expressed in a body fixed frame. Joint angles and velocities are represented by $\mathbf \theta$ and $\dot{\mathbf \theta}$, respectively.

For a straight-forward application of nonlinear optimal control to robotic systems, ct\_rbd offers wrapper classes which allow running nonlinear optimal control algorithms for any rigid-body dynamics model. 
As an easy way to handle contacts on arbitrary reference frames, we currently support a soft spring-damper contact model, which is described in detail in~\cite{giftthaler2017autodiff}.
Alternatively, contact forces can be chosen as additional control inputs. Furthermore, the CT allows to
\begin{itemize}
    \item generate operational-space models from the generated dynamics equations,
    \item augment rigid-body dynamic systems with arbitrary user-defined actuator dynamics models,
    \item use a number of pre-defined standard controllers such as joint position controllers plus inverse dynamics,
    \item use predefined cost function terms that are specific to robotic systems, e.g. we define auto-differentiable cost function terms for end-effector task-space positioning.
\end{itemize}
Lastly, we provide an interface for solving inverse kinematics problems using IKFast~\cite{ikfast}.

\section{Models Module}
\label{models_module}
The models module, ct\_models, contains a collection of fix- and floating base robot models which serve as examples of how to include systems in different ways:
\begin{itemize}
    \item the quadrotor is a floating-base system which is modeled independent from ct\_rbd and can serve as an example of how to implement a system which derives directly from \emph{core::ControlledSystem}.
    \item the inverted pendulum is the simplest system to be modeled using RobCoGen: a fix-base robot with 1~DoF.
    \item `HyA' models the fix-base,~6 DoF robot arm from~\cite{phd16brehman}.
    \item `HyQ'~\cite{semini:2011:hyqjournal} is a quadrupedal robot with 18~DoF.
    \item the quadrotor with slung-load is modeled with RobCoGen. It is an example of how to adapt \emph{rbd::Floating\-BaseFDSystem} for robots with unusual actuation.
\end{itemize}

For systems modeled using RobCoGen, ct\_models contains the generated dynamics code. ct\_models also gives examples of how to compile derivative code for forward and inverse dynamics into a separately loadable library.

\section{Application Examples}
\label{sec:application_examples}
The CT has been validated in a number of projects, including many hardware experiments, demonstrations, and academic publications. The following presents a compact summary. For details, we refer the interested reader to the referenced papers. CT application examples include
\begin{itemize}
\item NMPC for a hexrotor flying through a window\footnote{\url{https://youtu.be/Y7-1CBqs4x4}}~\cite{neunert16hexrotor},
\item a quadrotor with rotor failure performing a go-to task\footnote{\url{https://youtu.be/5MbnM2FiJ0M}},
\item Trajectory Optimization and full-body Nonlinear Model Predictive Control on different quadruped robots, including performing agile squat jumps\footnote{\url{https://youtu.be/vuCSKtP67E4}}~\cite{neunert:2017:ral,neunert2017mpc},
\item online trajectory optimization with collision avoidance \cite{giftthaler2017autodiff} on a 6~DoF industrial robot arm,
\item the computation of derivatives of constraints and cost functions including complex kinematic chains was demonstrated in hardware experiments in \footnote{\url{https://youtu.be/rVu1L_tPCoM}}~\cite{giftthaler2017efficient}.
\item pick-and-place arm motions for mobile manipulator were demonstrated in~\cite{sandy2016autonomous}.
\end{itemize}

In many of the above examples, our solvers reason about full rigid body dynamics models which are not simplified or altered by heuristics. Even for the most complex systems, the quadrupedal robots with 36~states and 12~control inputs, we can run our solvers in nonlinear MPC-fashion at rates higher than 150~Hz. These frequencies can be achieved even for long time horizons over 500~ms and for complicated locomotion tasks without pre-specified contact sequences, locations or timings.

\section{Performance Optimization}
\label{sec:performance_opt}
The Control Toolbox is optimized for performance and, if used correctly, constitutes one of the fastest implementations for many state-of-the-art control approaches. This section gives an outline of important steps to achieve the best performance.
To achieve best runtime performance, the CT can make of two main techniques:

\subsubsection{Multithreading}
Thorough multithreading can increase the performance of many optimal control algorithms. While some parts of the optimal control algorithms in the CT are strictly sequential (for instance the backward propagation of the Riccati equations), other parts can be entirely parallelized (e.g. the forward integration on separate multiple shooting intervals in DMS and GNMS and computing linear-quadratic approximations about solution candidates). When employing multi-threading, the required computation time decreases approximately linearly with the number of available cores. 
In practice, a trade-off needs to be achieved between single-core computation power (for the sequential algorithmic parts) and the overall number of cores (for the simultaneous parts). 
For experimental results on performance gains through multi-threading, the interested reader is referred to~\cite{giftthaler2017autodiff}.

\subsubsection{Vectorization}
To achieve the best runtime in every core, one can employ the processor's vectorization capabilities, which are Single Instruction Multiple Data (SIMD) implementations. SIMD is well-known to be particularly profitable an efficient execution of linear algebra operations, such as matrix-vector multiplications.
To date, the authors recommend employing AVX instructions~\cite{firasta2008intel}, as the register sizes of AVX are continuously growing in modern CPUs.

\section{Further Information}
\label{sec:further_information}
\small
The Control Toolbox is released under the Apache Licence, version 2.0.
More detailed documentation and a tutorial are available online, \url{https://adrlab.bitbucket.io/ct}. The source-code is available at \url{https://bitbucket.org/adrlab/ct}.

\section{Acknowledgements}
\label{sec:contributors}
\small
Developing and maintaining a large software framework is a team effort. We gratefully acknowledge the contributions of our current and former colleagues Farbod Farshidian, Diego Pardo, Timothy Sandy, Jan Carius, Ruben Grandia and Hamza Merzic.

This research has been funded through a Swiss National Science Foundation Professorship award to Jonas Buchli, the NCCR Digital Fabrication and the NCCR Robotics.

%\section{Funding}
%\small

\bibliographystyle{ieeetr}
\bibliography{root}

\begin{thebibliography}{10}

\bibitem{tedrake:2014:drake}
R.~Tedrake, ``Drake: A planning, control, and analysis toolbox for nonlinear
  dynamical systems.'' \url{http://drake.mit.edu}, 2014.
\newblock retrieved: 01-01-2018.

\bibitem{todorov2012mujoco}
E.~Todorov, T.~Erez, and Y.~Tassa, ``{MuJoCo: A physics engine for model-based
  control},'' in {\em IEEE/RSJ International Conference on Intelligent Robots
  and Systems (IROS)}, pp.~5026--5033, 2012.

\bibitem{houska2011acado}
B.~Houska, H.~J. Ferreau, and M.~Diehl, ``{ACADO toolkit-An open-source
  framework for automatic control and dynamic optimization},'' {\em Optimal
  Control Applications and Methods}, vol.~32, no.~3, pp.~298--312, 2011.

\bibitem{acados}
ACADOS, ``{ACADOS: Fast and Embedded Optimal Control Problem Solver}.''
  \url{https://github.com/acados/acados}.
\newblock retrieved: 2018-03-25.

\bibitem{becerra2010psopt}
V.~M. Becerra, ``Solving complex optimal control problems at no cost with
  psopt,'' in {\em Computer-Aided Control System Design (CACSD), 2010 IEEE
  International Symposium on}, pp.~1391--1396, IEEE, 2010.

\bibitem{diehl2001muscod}
M.~Diehl, D.~Leineweber, and A.~Sch{\"{a}}fer, ``{MUSCOD-II users' manual},''
  tech. rep., Interdisciplinary Center for Scientific Computing (IWR),
  University of Heidelberg, Germany, 2001.

\bibitem{patterson2014gpops}
M.~A. Patterson and A.~V. Rao, ``Gpops-ii: A matlab software for solving
  multiple-phase optimal control problems using hp-adaptive gaussian quadrature
  collocation methods and sparse nonlinear programming,'' {\em ACM Trans. Math.
  Softw.}, vol.~41, pp.~1:1--1:37, Oct. 2014.

\bibitem{forcespro}
{Embotech}, ``{FORCES Pro}.'' https://www.embotech.com/FORCES-Pro.
\newblock retrieved: 2018-01-01.

\bibitem{frisonalgorithms}
G.~Frison, {\em {Algorithms and Methods for Fast Model Predictive Control}}.
\newblock PhD thesis, Technical University of Denmark, 2015.

\bibitem{Ferreau2014qpoases}
H.~Ferreau, C.~Kirches, A.~Potschka, H.~Bock, and M.~Diehl, ``{qpOASES}: A
  parametric active-set algorithm for quadratic programming,'' {\em
  Mathematical Programming Computation}, vol.~6, no.~4, pp.~327--363, 2014.

\bibitem{wachter:2006:ipopt}
A.~W{\"a}chter and L.~T. Biegler, ``On the implementation of an interior-point
  filter line-search algorithm for large-scale nonlinear programming,'' {\em
  Mathematical programming}, vol.~106, no.~1, pp.~25--57, 2006.

\bibitem{gill:2005:snopt}
P.~E. Gill, W.~Murray, and M.~A. Saunders, ``{SNOPT}: An {SQP} algorithm for
  large-scale constrained optimization,'' {\em SIAM review}, vol.~47, no.~1,
  pp.~99--131, 2005.

\bibitem{nlopt}
S.~G. Johnson, ``{The NLopt nonlinear-optimization package}.''
  \url{http://ab-initio.mit.edu/nlopt}.
\newblock retrieved: 2018-01-01.

\bibitem{todorov2005ilqg}
E.~Todorov and W.~Li, ``A generalized iterative {LQG} method for
  locally-optimal feedback control of constrained nonlinear stochastic
  systems,'' in {\em American Control Conference, 2005. Proceedings of the
  2005}, pp.~300--306, IEEE, 2005.

\bibitem{giftthaler2017family}
M.~Giftthaler, M.~Neunert, M.~St{\"a}uble, J.~Buchli, and M.~Diehl, ``{A Family
  of Iterative Gauss-Newton Shooting Methods for Nonlinear Optimal Control}.''
  {arXiv:1711.11006 [cs.SY]}, 2017.

\bibitem{bell2012cppad}
B.~M. Bell, ``{CppAD: a package for C++ algorithmic differentiation},'' {\em
  Computational Infrastructure for Operations Research}, vol.~57, 2012.

\bibitem{web:matlab}
{\relax MathWorks Inc.}, ``{Matlab \& Simulink software}.''
  \url{www.mathworks.com/}, 2017.
\newblock [Online, retrieved: 2017-12-23].

\bibitem{web:mathematica}
{\relax Wolfram Research}, ``{Mathematica Software}.'' \url{www.wolfram.com/},
  2018.
\newblock [Online, retrieved: 2018-01-03].

\bibitem{web:maplesoft}
{\relax Maplesim}, ``{Maplesoft}.''
  \url{https://www.maplesoft.com/products/maplesim/}, 2018.
\newblock [Online, retrieved: 2018-03-25].

\bibitem{giftthaler2017autodiff}
M.~Giftthaler, M.~Neunert, M.~St\"auble, M.~Frigerio, C.~Semini, and J.~Buchli,
  ``Automatic differentiation of rigid body dynamics for optimal control and
  estimation,'' {\em Advanced Robotics}, November 2017.

\bibitem{felis2016rbdl}
M.~L. Felis, ``{RBDL: an efficient rigid-body dynamics library using recursive
  algorithms},'' {\em Autonomous Robots}, vol.~14, no.~2, p.~495–511, 2016.

\bibitem{dart}
K.~Liu, ``{Dynamic Animation and Robotics Toolkit (DART)}.''
  \url{http://dartsim.github.io}.
\newblock [Online, retrieved 02-02-2018].

\bibitem{pinocchioweb}
J.~Carpentier, F.~Valenza, N.~Mansard, {\em et~al.}, ``Pinocchio: fast forward
  and inverse dynamics for poly-articulated systems.''
  \url{https://stack-of-tasks.github.io/pinocchio}, 2015--2018.

\bibitem{idyntree}
F.~Nori, S.~Traversaro, J.~Eljaik, F.~Romano, A.~Del~Prete, and D.~Pucci,
  ``icub whole-body control through force regulation on rigid noncoplanar
  contacts,'' {\em Frontiers in Robotics and AI}, vol.~2, no.~6, 2015.

\bibitem{frigerio:2016:robcogen}
M.~Frigerio, J.~Buchli, D.~G. Caldwell, and C.~Semini, ``{R}ob{C}o{G}en: a code
  generator for efficient kinematics and dynamics of articulated robots, based
  on {D}omain {S}pecific {L}anguages,'' {\em Journal of Software Engineering
  for Robotics (JOSER)}, vol.~7, no.~1, pp.~36--54, 2016.

\bibitem{pardo2017projection}
D.~Pardo, M.~Neunert, A.~Winkler, R.~Grandia, and J.~Buchli, ``Hybrid direct
  collocation and control in the constraint-consistent subspace for dynamic
  legged robot locomotion,'' in {\em Proceedings of Robotics: Science and
  Systems}, (Cambridge, Massachusetts), July 2017.

\bibitem{web:eigen}
{E}igen Linear Algebra~Library, 2017.
\newblock eigen.tuxfamily.org, retrieved: 2017-08-25.

\bibitem{kindr}
C.~D. Bellicoso, M.~Bl\"osch, R.~Diethelm, P.~Fankhauser, P.~Furgale,
  C.~Gehring, and H.~Sommer, ``{Kindr - Kinematics and Dynamics for
  Robotics}.'' \url{https://docs.leggedrobotics.com/kindr/index.html}.
\newblock {accessed 01-01-2018}.

\bibitem{odeint}
K.~Ahnert and M.~Mulansky, ``{Odeint – Solving Ordinary Differential
  Equations in C++},'' in {\em AIP Conference Proceedings}, pp.~1586--1589,
  2011.

\bibitem{sherman2011simbody}
M.~A. Sherman, A.~Seth, and S.~L. Delp, ``Simbody: multibody dynamics for
  biomedical research,'' {\em Procedia IUTAM}, vol.~2, pp.~241 -- 261, 2011.
\newblock IUTAM Symposium on Human Body Dynamics.

\bibitem{mayne1966ddp}
D.~Mayne, ``A second-order gradient method for determining optimal trajectories
  of non-linear discrete-time systems,'' {\em International Journal of
  Control}, vol.~3, no.~1, pp.~85--95, 1966.

\bibitem{blasfeo}
G.~Frison, D.~Kouzoupis, T.~Sartor, A.~Zanelli, and M.~Diehl, ``{BLASFEO: basic
  linear algebra subroutines for embedded optimization},'' 2017.
\newblock {arXiv:1704.02457 [cs.MS]}.

\bibitem{bock1984direct}
H.~Bock and K.~Plitt, ``A multiple shooting algorithm for direct solution of
  optimal control problems,'' {\em IFAC World Congress Proceedings Volumes},
  vol.~17, no.~2, pp.~1603 -- 1608, 1984.

\bibitem{nocedal}
J.~Nocedal and S.~J. Wright, {\em {Numerical Optimization}}.
\newblock Springer, 1999.

\bibitem{diehl2005rti}
M.~Diehl, R.~Findeisen, F.~Allgower, H.~G. Bock, and J.~P. Schloder, ``Nominal
  stability of real-time iteration scheme for nonlinear model predictive
  control,'' {\em IEE Proceedings - Control Theory and Applications}, vol.~152,
  pp.~296--308, May 2005.

\bibitem{neunert2017mpc}
M.~Neunert, M.~St{\"a}uble, M.~Giftthaler, C.~D. Bellicoso, J.~Carius,
  C.~Gehring, M.~Hutter, and J.~Buchli, ``{Whole-Body Nonlinear Model
  Predictive Control Through Contacts for Quadrupeds},'' 2018.
\newblock {IEEE Robotics and Automation Letters}.

\bibitem{ikfast}
R.~Diankov, ``{ Ikfast: The robot kinematics compiler}.''
  \url{http://openrave.org/docs/0.8.2/openravepy/ikfast/}.
\newblock retrieved: 2018-01-01.

\bibitem{phd16brehman}
B.~U. Rehman, {\em Design and Control of a Compact Hydraulic Manipulator for
  Quadruped Robots}.
\newblock PhD thesis, Istituto Italiano di Tecnologia (IIT) and University of
  Genova, 2016.

\bibitem{semini:2011:hyqjournal}
C.~Semini, N.~G. Tsagarakis, E.~Guglielmino, M.~Focchi, F.~Cannella, and D.~G.
  Caldwell, ``Design of {HyQ}--a hydraulically and electrically actuated
  quadruped robot,'' {\em Institution of Mechanical Engineers, Journal of
  Systems and Control Engineering}, vol.~225, pp.~831--849, 2011.

\bibitem{neunert16hexrotor}
M.~Neunert, C.~de~Crousaz, F.~Furrer, M.~Kamel, F.~Farshidian, R.~Siegwart, and
  J.~Buchli, ``Fast nonlinear model predictive control for unified trajectory
  optimization and tracking,'' in {\em IEEE International Conference on
  Robotics and Automation}, 2016.

\bibitem{neunert:2017:ral}
M.~Neunert, F.~Farshidian, A.~W. Winkler, and J.~Buchli, ``Trajectory
  optimization through contacts and automatic gait discovery for quadrupeds,''
  {\em IEEE Robotics and Automation Letters (RA-L)}, 2017.

\bibitem{giftthaler2017efficient}
M.~Giftthaler, F.~Farshidian, T.~Sandy, L.~Stadelmann, and J.~Buchli,
  ``{Efficient Kinematic Planning for Mobile Manipulators with Non-holonomic
  Constraints Using Optimal Control},'' in {\em IEEE International Conference
  on Robotics and Automation}, pp.~3411--3417, May 2017.

\bibitem{sandy2016autonomous}
T.~Sandy, M.~Giftthaler, K.~D\"orfler, M.~Kohler, and J.~Buchli, ``{Autonomous
  repositioning and localization of an In Situ Fabricator},'' in {\em IEEE
  International Conference on Robotics and Automation}, pp.~2852--2858, May
  2016.

\bibitem{firasta2008intel}
N.~Firasta, M.~Buxton, P.~Jinbo, K.~Nasri, and S.~Kuo, ``Intel avx: New
  frontiers in performance improvements and energy efficiency,'' {\em Intel
  white paper}, vol.~19, p.~20, 2008.

\end{thebibliography}

\end{document}